# Deep Learning: a Heuristic Three-stage Mechanism for Grid Searches to Optimize the Future Risk Prediction of Breast Cancer Metastasis Using EHR-based Clinical Data

Xia Jiang[1], Yijun Zhou[1], Chuhan Xu[1], Adam Brufsky[3,4], Alan Wells[2,4]

[1]Department of Biomedical Informatics
University of Pittsburgh
Pittsburgh, PA

[2]Department of Pathology
University of Pittsburgh and Pittsburgh VA Health System
Pittsburgh, PA

[3]Division of Hematology/Oncology
University of Pittsburgh School of Medicine
Pittsburgh, PA

[4]UPMC Hillman Cancer Center
Pittsburgh, PA

**Contact:**       Xia Jiang

**Email:**         xij6@pitt.edu

**Phone:**         412-648-9310

Yijun Zhou: yiz209@pitt.edu
Chuhan Xu: xuchuhan777@gmail.com
Alan Wells: wellsa@upmc.edu
Adam Brufsky: brufskyam@upmc.edu

# ABSTRACT

**Background**
A grid search, at the cost of training and testing a large number of models, is an effective way to optimize the prediction performance of deep learning models. A challenging task concerning grid search is the time management. Without a good time management scheme, a grid search can easily be set off as a "mission" that will not finish in our lifetime. In this study, we introduce a heuristic three-stage mechanism for managing the running time of low-budget grid searches with deep learning, and the *sweet-spot grid search* (*SSGS*) and *randomized grid search* (*RGS*) strategies for improving model prediction performance, in an application of predicting the 5-year, 10-year, and 15-year risk of breast cancer metastasis.

**Methods**
We *develop deep feedforward neural network* (*DFNN*) models and optimize the prediction performance of these models through grid searches. We conduct eight cycles of grid searches in three stages, focusing on learning a reasonable range of values for each of the adjustable hyperparameters in Stage 1, learning the sweet-spot values of the set of hyperparameters and estimating the unit grid search time in Stage 2, and conducting multiple cycles of timed grid searches to refine model prediction performance with SSGS and RGS in Stage 3. We conduct various SHAP analyses to explain the prediction, including a unique type of SHAP analyses to interpret the contributions of the DFNN-model hyperparameters.

**Results**
The grid searches we conducted improved the risk prediction of 5-year, 10-year, and 15-year breast cancer metastasis by 18.6%, 16.3%, and 17.3% respectively, over the average performance of all corresponding models we trained using the RGS strategy.

**Conclusions**
Grid search can greatly improve model prediction. Our result analyses not only demonstrate best model performance but also characterize grid searches from various aspects such as their capabilities of discovering decent models and the unit grid search time. The three-stage mechanism worked effectively. It not only made our low-budget grid searches feasible and manageable, but also helped improve the model prediction performance of the DFNN models. Our SHAP analyses not only identified clinical risk factors important for the prediction of future risk of breast cancer metastasis, but also DFNN-model hyperparameters important to the prediction of performance scores.

**Keywords:** deep learning, machine learning, grid search, neural networks, breast cancer metastasis, metastatic breast cancer, breast cancer, metastasis, prediction, EHR, clinical

# INTRODUCTION

Electronic health records (EHR) systems have been used in clinical settings for many years. With the help of these systems, researchers or medical practitioners who work in a clinical environment often have opportunities to access and curate a clinical dataset concerning a group of patients, and sometimes they may consider using such a dataset to build a patient outcome prediction model. For instance, one can use the EHR data collected from a breast cancer patient care center to develop deep learning models that can be used to predict for a patient the risk of future occurrences of breast cancer metastasis.

Breast cancer is a major cancer related cause of death for women worldwide. Based on the report updated on March 13, 2024 by the World Health Organization (WHO), "Breast cancer was also the most common cancer in women in 157 countries out of 185 in 2022" , and it is also one of the main causes of cancer related death in women worldwide, and "caused 670,000 deaths globally in 2022" . Again according to the WHO, as of the end of 2020, "there were 7.8 million women alive who were diagnosed with breast cancer in the past 5 years, making it the world's most prevalent cancer". Women do not die of breast cancer, rather, they die mainly due to breast cancer metastasis, which can occur years after the initial treatment of breast cancer [1,2]. Predicting a late metastatic occurrence of breast cancer for a patient is important, because the prediction can help making more suitable treatment plan for the patient, which may help prevent breast cancer metastasis. Improving our capability of predicting breast cancer metastasis is an important task in breast cancer patient care. Even just a small percentage of improvement can help greatly improve patient quality of life, and save lives and care related costs.

The field of machine learning (ML) and deep learning [3–8] has provided us with various AI-based computational methods for conducting predictions. Using these ML methods we can learn a prediction model automatically from a dataset. However, a prediction model that is developed in such a manner does not always predict well [9]. There are often multiple factors that can affect model performance. For example, the model performance is usually dataset dependent, that is, the same machine learning method can perform totally differently when it is applied to different datasets [4,9,10]. This phenomenon is perhaps partly because different datasets contain different levels of "signals" that are critical in making correct predictions. When a dataset contains very weak signals, even an advanced method can fail learning a good prediction model. By "signals" we mean the information usable for making predictions, contained in data. A good example of a signal is what so-called a correlation between two variables. If two variables are correlated, then one of them can be used to predict the other, and in that case, the former is often called the predictor and the latter is often called the outcome. Sometimes, to curate a dataset that contains sufficient information for learning a good prediction model, we need to collect a lot of data. Generally speaking, the more datapoints (cases) a dataset contains, the more likely it provides sufficient information for learning a good prediction model. That perhaps explains in an aspect why the applications of ML methods are often associated with the term "big data" and a field called data science.

Other than the dataset itself, another important factor for model performance is the value used for an adjustable hyperparameter of a prediction model. All machine learning methods that we used so far have adjustable hyperparameters, but a difference is some methods have more and some have less [4,11–14]. During the early years of using machine learning methods to carry on real-life prediction tasks, we paid little attention to the selection of a value for an adjustable hyperparameter that is built into a machine learning method [9,10,15]. A normal practice is to use the default value recommended by the developers of the machine learning method or by a machine learning textbook, or at most try a few values that are close to the default value. This may have contributed significantly to the fact that some of the ML method such as the first generation of the Neural Network were reportedly having poor prediction performance [9,10,16].

Deep learning is a machine learning method that has quite some adjustable hyperparameters [17–19]. For example, we identified 13 adjustable hyperparameters (see Table 1 and a more detailed description of these hyperparameters in Table 3) for the Deep Feedforward Neural Network (DFNN) models that we developed for predicting later occurrences of breast cancer metastasis [4,19]. In a previous study, by conducting machine learning experiments, we found that different value assignments of the adjustable hyperparameters can lead to models with significantly different prediction performance [4]. We used a method called *grid search* [4,20,21] to systematically train and test different DFNN models by changing the values of the set of adjustable hyperparameters [4]. Therefore, in order to do a grid search we normally preselect a range of ranges for each of the hyperparameters as an input to the grid search.

We now use an example to explain what a grid search does and in the meantime introduce a major challenge of conducting a grid search. In this example, we use the number of values given to each of the 13 hyperparameters as shown in the second row of Table 1 below. The number of hidden layers and the number of hidden nodes (neurons) in a hidden layer are the two structural hyperparameters that together determines the structure of a DFNN model. In this example grid search, the number of hidden layers has four different values, that is, the model can contain up to four hidden layers. The number of hidden nodes per hidden layer has 22 different values. Therefore, there could be 22, 484, 10648, or 234,256 different model structures when the number of hidden layer is configured to be 1, 2, 3, or 4. Thus, we can make in total 245,410 different models by considering the two structural hyperparameters alone. The total number of possible unique value assignments to the set of the 11 remaining non-structural hyperparameters are the product of the number of values given to each of these hyperparameters, which arrives at 1.7424e9. Therefore, considering all 13 hyperparameters, there are in total 4.276e14 unique value assignments. What the example grid search does is to train and test 4.276e14 DFNN models determined by the 4.276e14 different value assignments, one at a time. We call a unique value assignment to the set of hyperparameters of a grid search a *hyperparameter setting*. Note that under each of the hyperparameter settings, there would be k different models trained and tested if the k-fold cross validation (CV) procedure (see the Methods section) is applied. Since we use a 5-fold CV procedure in our grid searches, the number of models trained and tested is five times the number of hyperparameters settings used in these grid searches.

Table 1: The number of values for each of the adjustable hyperparameters of a DFNN model, used in the example grid search

*NHL: number of hidden layers; NHN: number of hidden nodes; AF: activation function; KI: kernel initializer; O: optimizer; LR: learning rate; M: momentum; ID: iteration-based decay; DR: dropout rate; E: epochs; BS: batch_size.

| Hyperparameter Name Acronym* | NHL | NHN | AF | FI | O | LR | M | ID | DR | E | BS | L1 | L2 |
|---|---|---|---|---|---|---|---|---|---|---|---|---|---|
| Number of Values | 4 | 22 | 4 | 5 | 5 | 10 | 4 | 11 | 3 | 12 | 11 | 10 | 10 |

A grid search can be very costly! Based on the grade search experiments we did in our previous study [4], the average running time per hyperparameter setting is 117 seconds for a particular dataset we used. By using this average unit running time, the estimated total running time for the example grid search, as described above, would be 1,586,424,369 years. Apparently this grid search is not feasible for us unless we use billions of computers to run it parallelly. A grid search is in general very time consuming, but computation time can sometimes be resolved by using high-speed computing, which can be bought by money. Therefore, the feasibility issue of a grid search essentially boils down to a financial budget issue. Going back to the scenario that we mentioned in the beginning of this introduction, in which medical researchers or practitioners want to learn a prediction model using their own EHR-based datasets, a normal situation is that the time and funds available are both very limited for conducting a grid search to optimize prediction. We call a grid search for which the money allocation is very limited a low-budget grid search. It is not uncommon to encounter a low-budget grid search in the real-life applications of machine learning methods.

Moreover, the example grid search we described above can only be called a small-scale grid search, in which only a small number of values is given to each of the hyperparameters. But the estimated running time is already hard to manage. Based on this example, we see that the number of values allowed for each of the hyperparameters can only be very small for a low-budget grid search to finish in a foreseeable time. But note that some of the hyperparameters of our DFNN models can take a very large number of different values (see Table 3). For example, based on our previous studies [4,19], the range of values we consider for a hyperparameter called *epochs* is from 5 to 2000, which means for each grid search we need to select a very small set of values from the 1996 different values for epochs. As a matter of fact, most of the 13 hyperparameters can take a very large number of values, and some can even take an infinite number of different values within a normal range. It is often a challenging task to select a very small set of values from a large number of values available for each of the hyperparameters like epochs, and in the meantime to ensure the feasibility of a low-budget grid and meet the goal of digging out a better prediction model through the grid search. To our knowledge, there is no standard and good way of doing this! In this study, we have limited time to run grid searches for optimizing our DFNN models that predict the risk of breast cancer metastasis. We therefore introduce what we call a three-stage heuristic grid-search mechanism that we use to manage this challenging task. We describe this mechanism and the experiments we conduct, in which we apply this mechanism, in the Methods section below, and present and analyze the results of the experiments in the Results and Discussion sections.

## METHODS

### The DFNN Models

**Deep learning and deep feedforward neural network (DFNN):** An *Artificial Neural Network* (*ANN*) is a machine learning framework, which is designed to recognize patterns using a model loosely resembling the human brain [22,23]. ANNs can be used for clustering (unsupervised) on unlabeled data or classification (supervised) on labeled data [4]. *Deep Neural Networks* (*DNNs*), called *deep learning*, refers to the use of neural networks composed of more than one hidden layers [3,5–7,24]. The DNN has obtained significant success in commercial applications such as voice and pattern recognition, computer vision, and image-based processing [25–35]. However, its power has not been fully explored or demonstrated in applications that are not image-based, such as the prediction of breast cancer metastasis using non-image clinical data. This is due in part to the sheer magnitude of the number of variables involved in these problems, which presents formidable computational and modeling challenges [17,18,36]. We developed the *deep feedforward neural network (DFNN)* models that predict the risk of a future occurrence of breast cancer metastasis for a patient [4,19] and conduct grid searches to hand these challenges. The DFNN models we developed are fully connected neural networks that do not contain cycles. Figure 1 illustrates, as an example, the structure and the inner connections of the DFNNs that we have developed. The example showed in Figure 1 is a six-layer neural network that contains one input layer, four hidden layers, and one output layer. The 31 input nodes to this neural network represent the 31 clinical features contained in the patient data that we use, and the output layer contains two nodes representing the binary status of 5-year, 10-year, or 15-year breast cancer metastasis. Each node in this model has an *activation function* (see Table 3), represented by *f(x)*, which decides the node's individual output value established by the current value of the node. In such a DFNN model, each hidden layer has a certain number of hidden nodes that can be different from the other layers. Both the number of hidden layers and the number of hidden nodes in a hidden layer are two of the set of hyperparameters whose values are subjected to changes during grid searches. These two hyperparameters together with other adjustable hyperparameters for our grid searchers are described in Table 3 below.

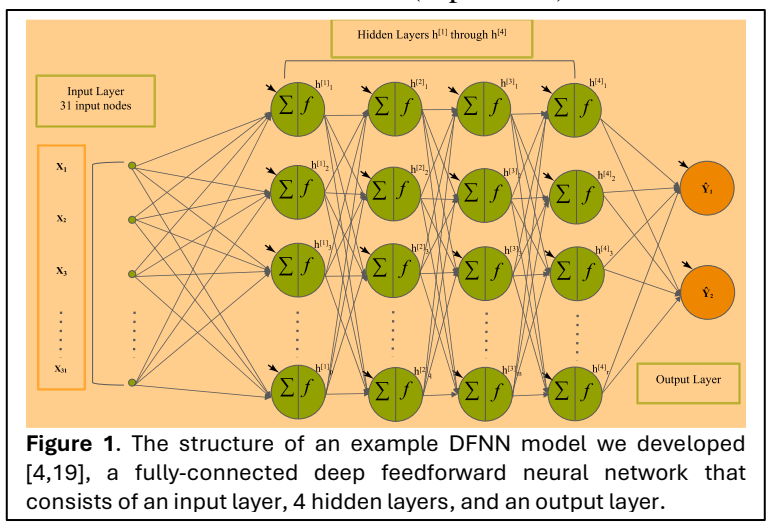

**Figure 1**. The structure of an example DFNN model we developed [4,19], a fully-connected deep feedforward neural network that consists of an input layer, 4 hidden layers, and an output layer.

**Datasets**

The MBIL method is a Bayesian Network-based method for identifying risk factors (RFs) for an outcome feature, which was applied to three EHR-based clinical datasets concerning breast cancer metastasis, that is, the LSM-5Year, 10Year, and 15year datasets [37]. In this study, we use MBIL to retrieve all RFs concerning breast cancer metastasis from the LSM datasets and develop three new datasets according to the RFs: the LSM_RF-5Year, 10Year, and 15Year datasets. Using the LSM_RF-5Year dataset as an example, the 2-step procedure for developing this dataset is as follows: Step 1: Applying the MBIL method to the LSM-5Year dataset to retrieve the RFs of 5-year breast cancer metastasis. The original LSM-5Year dataset contains 32 features including a feature called metastasis, which represents the state of having or not having breast cancer metastasis by the $5^{th}$ year post the initial treatment [37]. In this study metastasis is the outcome feature, which we also call the target feature, because we are interested in predicting the value of this features using the other features; The remaining 31 features are called predictors. Step 2: Removing all predictors that don't belong to the set of RFs found in Step 1 from the LSM-5Year dataset, and all data points of the remaining features form the LSM_RF-5Year dataset. We also follow this 2-step procedure to obtain the LSM_RF-10Year dataset from the original LSM-10Year dataset and the LSM_RF-15Year dataset from the LSM-15Year dataset. All experiments conducted in this study are based on the RF datasets. Table 2 below shows the counts of the cases and predictors included in the three RF datasets. A detailed description of the predictors is included in the Tables S1-S3 of the supplement.

**Table 2:** Case and predictor counts of the three LSM_RF datasets

|  | Total # of cases | # Positive cases | # Negative cases | # of Predictors |
|---|---|---|---|---|
| LSM-RF-5year | 4189 | 437 | 3752 | 20 |
| LSM-RF-10year | 1827 | 572 | 1255 | 18 |
| LSM-RF-15year | 751 | 608 | 143 | 17 |

# The three-stage grid search mechanism and the Experiments

In this study, we tune 13 hyperparameters (Table 3) following a heuristic three-stage grid search mechanism for learning the DFNN models. In Stage 1, we focus on learning performance trend and a set of proper values for each of the hyperparameters that can take a very large or an infinite number of different values, such as the epochs and learning rate. The extreme values that can result in performance outliers, that is, the models that perform very poorly, are usually removed from the set of proper values. Stage 2 and stage 3 grid searches are guided by the set of proper values. In Stage 2, we attempt to estimate the average *running time per hyperparameter setting* (RTPS) and identify performance sweet spots by conducting a grid search that is guided by the results of the Stage 1 grid searches. A main difference between the Stage 1 and the Stage 2 grid searches is that in a Stage 1 grid search we only change the values of a single hyperparameter, and this allows the Stage 1 grid search to train and test models using a large number of different values of the hyperparameter and still finish in an acceptable timeframe. But in a Stage 2 grid search or a Stage 3 grid search, we allow all hyperparameters to take multiple different values. The Stage 2 experiments are important because based on the results of such a grid search, we can do better in managing the running time of the subsequent Stage 3 grid searches. Specifically, we can estimate the total running time based on the RTPS if we know the input number of hyperparameter settings, and on the other hand, we can computer the input number of hyperparameter settings based the total running time allowed. This ensures that a grid search finishes within an acceptable timeframe, and therefore is critical to a low-budget and time-sensitive grid search. In Stage 3, we conduct multiple cycles of grid searches to further refining mode prediction performance. A cycle means the entire process of running a grid search from preselecting a set of values for each hyperparameter till finishing all experiments concerning model training and testing scheduled for the grid search. In this study, we conduct 6-cycles of Stage 3 grid searches for each of the three datasets, labeled as Stage3-c1, c2, and up to c6, each respectively. The hyperparameter settings of the stage3-c1 are determined based on the results of the Stage 2 grid searches, that is, when conducting hyperparameter value selection, we focus on a small number of values that are close to the value used by the best model obtained from the Stage 2 grid searches. We call such a grid search a *sweet spot grid search (SSGS)*, in which the sweet spot is determined by a previous grid search. Stage 3 cycle 1 through 5 are all SSGSs. In Stage 3, we also use our *out of the local optimal* (OLO) strategy for preselecting the input hyperparameter values. Specifically, in Stage3-c4 and Stage3-c5, we identify a new sweet spot that is outside the "neighborhood" of the previous sweet spot to get out of the potential local optimal formed during the SSGSs. Finally, in Stage3-c6, we apply both OLO and a strategy called the *randomized grid search (RGS)* that we created. Once the proper set of values for each of the hyperparameters is determined via Stage 1, we will be able to determine all possible unique hyperparameter settings, which we call the *pool of hyperparameter settings (PHS)*. A RGS is a grid search that trains and tests models at a set of hyperparameters settings that are randomly picked from a PHS. Table 3 below contains more detailed information about the 13 hyperparameters of our DFNN models. The specific range of values used in each cycle of our three-stage grid searches are shown in Table S4 to S6 of the supplement.

**Table 3.** The set of adjustable hyperparameters and their ranges of values considered for our grid searches

| Hyperparameter | Description | Values |
|---|---|---|
| Number of hidden layers | The depth of a DNN | 1,2,3,4 |
| Number of hidden nodes in a hidden layer | Number of neurons in a hidden layer | 1-1005 |
| Activation function | It determines the value to be passed to the next node based on the value of the current node | Relu, Sigmoid, Softmax, Tanh |
| Kernel initializer | Assigning initial values to the internal model parameters | Constant, Glorot_normal, Glorot_uniform, He_normal, He_uniform |
| Optimizer | Optimizes internal model parameters towards minimizing the loss | SGD, Adam, Adagrad, Nadam, Adamax |
| Learning rate | Used by both SGD and Adagrad | 0.001 ~ 0.3 |
| Momentum | Smooths out the curve of gradients by moving average. Used by SGD. | 0 ~ 0.9 |
| Iteration-based decay | Iteration-based decay; updating learning rate by a decreasing factor in each epoch | 0 ~ 0.1 |
| Dropout rate | Manage overfitting and training time by randomly selects nodes to ignore | 0 ~ 0.5 |
| Epochs | Number of times model is trained by each of the training set samples exactly one | 5 ~ 2000 |
| Batch_size | Unit number of samples fed to the optimizer before updating weights | 1 to the # of datapoints in a dataset |
| L1 | Sparsity regularization; | 0 ~ 0.03 |
| L2 | Weight decay regularization; it penalizes large weights to adjust the weight updating step | 0 ~ 0.2 |

**Prediction performance metrics and the 5-fold cross validation process**

Our grid searches follow the 5-fold cross validation (CV) mechanism to train and test models at each hyperparameter setting, and use an AUC score to measure the prediction performance of a model. AUC stands for the *area under the curve* of a *receiver operator characteristic* (ROC) curve that plots the true positive rate against the false positive rate for all possible cutoff values [38] . An AUC score measures the discrimination performance of the model.

*To conduct a 5-fold CV*, we need to split a dataset prior to grid searches. We use the following procedure to split our datasets: 1) split the entire dataset into a train-test set that contains 80% of the cases and a validation set that contains 20% of the cases. The train-test set will be given to a grid search as the input dataset, and the validation set will be kept aside for a later validation test. 2) divide a train-test set evenly into 5 portions for conducting the 5-fold CV. The division is mostly done randomly except that each portion should have approximately 20% of the positive cases and 20% of the negative cases to ensure that it is a representative fraction of the dataset. We conduct a 5-fold CV at each hyperparameter setting of a grid search. During a 5-fold CV, 5 different models are generated and tested, each is trained using a unique combination of 4 portions and tested with the remaining portion. 5 AUC scores are produced based on the tests, and the average of these scores is called the *mean_test_AUC*. The mean_test_AUC metric is used by grid searches to measure the model discrimination performance. A top hyperparameter setting selection at the end of a grid search is also based on this metric. A top model is developed by refitting the entire train-test set using the top hyperparameter setting selected by a grid search.

**The SHAP values and plots**

We use the SHAP (Shapley Additive Explanations) values to explain the prediction results of our DFNN models and identify the important features for the predicted future risk of breast cancer metastasis. The SHAP values are established based on the Shapley values, which distribute the payoff by measuring the marginal contribution of individual team members to the outcome of a cooperation game [39,40].

We use the Kernel Explainer provided by the SHAP package [41] to compute the SHAP values for the DFNN models. In order to manage computation time while preserving the integrity of the information contained in data, we process the training dataset using the k-means clustering method. We therefore identify k representative cluster centroids, where k is equal to the number of features in the dataset. These centroids, which epitomize the typical characteristics of the training dataset, are then employed as the background data for the SHAP Kernel Explainer. The *background value* of a feature is the mean of the corresponding feature values of the k centroids, identified using the k-means clustering.

The formula used to compute SHAP values is as follows:

$$\phi_i(p) = \sum S \subseteq F\setminus\{i\} \frac{|S|!(|F|-|S|-1)!}{|F|!} [p(S \cup \{i\}) - p(S)]$$

In this formula, $F$ denotes the complete set of features, $i$ represents the $i$th feature, and S represents a subset of $F$ not including the $i$th feature, that is, $S \subseteq F\setminus\{i\}$. $|S|$ is the size of a subset $S$, and $|F|$ is the size of the set $F$. Let $p(x)$ represents a model's prediction outcome, then $p(S \cup \{i\})$ represents the model's prediction output when using the features in a subset S together with the $i$th feature as the predictors, while $p(S)$ denotes the model's prediction output when using only the features in S as the predictors. The term $p(S \cup \{i\}) - p(S)$ therefore reflects the contribution to the model's prediction output made by the $i$th feature with respect to a subset S.

Inspired by LIME [42], the Kernel Explainer generates synthetic samples, which are used as the test cases for a model of interest to compute the SHAP value of a feature. Let's call the feature the $i$th feature. Each of the synthetic samples contains the real values of a subset S, taken from the 20% set-aside validation dataset (see Methods), and the background values for the remaining features in $F$, excluding the $i$th feature. The $i$th feature takes its real value for the synthetic sample that is created for obtaining the $p(S \cup \{i\})$, and takes its corresponding background value in the synthetic sample that is created to obtain the $p(S)$. Recall that the background values for all features in F are generated using the k-means method as described above.

We also use the SHAP package to generate a SHAP bar plot which ranks the feature importance values of all predictors from high to low. A SHAP feature importance value is the mean absolute SHAP value of all the test cases for the feature of interest. Another type of SHAP plots we show are the so called SHAP summary plots, which not only rank the features by their feature importance values, but also show the SHAP value of each individual case that is tested. The SHAP heatmap plots show patterns of groups of instances or features, and the dependence plots explore potential interactive features which jointly affect a SHAP value. .

# RESULTS

We followed the three-stage mechanism, described in the Methods section, and ran grid search experiments in eight cycles: 1 cycle for Stage 1, 1 cycle for stage 2, and 6 cycles for stage 3, named as Stage 3-c1, Stage 3-c2, and so on and so forth. For each cycle, we identified groups of top performing models, that is, the top 1, top 5, top 10, top 50, and top 100 models, out of all models that were trained. We then computed the average mean_test_AUC for each group as a measurement of group prediction performance. We also obtained the average mean_test_AUC for all models trained during a cycle of grid searches as the largest group. We compare the group prediction performance side by side of the eight cycles of grid searches for predicting 5-year breast cancer metastasis in Table 4, 10-year in Table 5, and 15-year in Table 6 below.

**Table 4.** Comparison of group prediction performance in average mean_test_AUCs for the DFNN-5Year models

| group | results | Stage1 | Stage2 | Stage3-c1 | Stage3-c2 | Stage3-3 | Stage3-c4 | Stage3-c5 | Stage3-c6 |
|---|---|---|---|---|---|---|---|---|---|
| Top1 | best | 0.75301 | 0.74812 | 0.75024 | 0.75066 | 0.75418 | 0.75386 | 0.75250 | 0.75825 |
| Top5 | Avg | 0.75225 | 0.74725 | 0.74894 | 0.74980 | 0.75344 | 0.75260 | 0.75219 | 0.75532 |
| Top 10 | Avg | 0.75144 | 0.74682 | 0.74658 | 0.74923 | 0.75274 | 0.75179 | 0.75194 | 0.75435 |
| Top 50 | Avg | 0.74852 | 0.7456 | 0.74658 | 0.74767 | 0.75114 | 0.75031 | 0.75091 | 0.75206 |
| Top 100 | Avg | 0.74692 | 0.7449 | 0.74584 | 0.74704 | 0.75044 | 0.74956 | 0.75038 | 0.75122 |
| All | Avg | 0.66456 | 0.6839 | 0.72096 | 0.73505 | 0.73657 | 0.72052 | 0.71997 | 0.63960 |

**Table 5**. Comparison of group prediction performance in average mean_test_AUCs for the DFNN-10Year models

| group | results | Stage1 | Stage2 | Stage3-c1 | Stage3-c2 | Stage3-c3 | Stage3-c4 | Stage3-c5 | Stage3-c6 |
|---|---|---|---|---|---|---|---|---|---|
| Top1 | best | 0.77789 | 0.78280 | 0.78802 | 0.79023 | 0.79152 | 0.78885 | 0.78059 | 0.78564 |
| Top5 | Avg | 0.77593 | 0.78141 | 0.78696 | 0.78897 | 0.79030 | 0.78762 | 0.77936 | 0.78272 |
| Top10 | Avg | 0.77511 | 0.78090 | 0.78626 | 0.78820 | 0.78980 | 0.78703 | 0.77889 | 0.78189 |
| Top50 | Avg | 0.77348 | 0.77942 | 0.78475 | 0.78644 | 0.78865 | 0.78553 | 0.77731 | 0.77992 |
| Top100 | Avg | 0.77257 | 0.77866 | 0.78402 | 0.78551 | 0.78797 | 0.78483 | 0.77652 | 0.77900 |
| All | Avg | 0.71344 | 0.73360 | 0.76124 | 0.75901 | 0.76702 | 0.76194 | 0.74336 | 0.68075 |

**Table 6:** Comparison of group prediction performance in average mean_test_AUCs for the DFNN-15Year models

| group | results | Stage1 | Stage2 | Stage3-c1 | Stage3-c2 | Stage3-c3 | Stage3-c4 | Stage3-c5 | Stage3-c6 |
|---|---|---|---|---|---|---|---|---|---|
| Top1 | best | 0.86502 | 0.86548 | 0.88023 | 0.87463 | 0.87359 | 0.87551 | 0.87255 | 0.86657 |
| Top5 | Avg | 0.86232 | 0.86403 | 0.87632 | 0.87397 | 0.87337 | 0.87476 | 0.87095 | 0.86578 |
| Top 10 | Avg | 0.86105 | 0.86294 | 0.87426 | 0.87353 | 0.87280 | 0.87427 | 0.87028 | 0.86510 |
| Top 50 | Avg | 0.85806 | 0.86069 | 0.87073 | 0.87149 | 0.87082 | 0.87281 | 0.86872 | 0.86271 |
| Top 100 | Avg | 0.85678 | 0.85966 | 0.86950 | 0.87055 | 0.86954 | 0.87199 | 0.86789 | 0.86170 |
| All | Avg | 0.76512 | 0.78853 | 0.83167 | 0.84535 | 0.83399 | 0.83782 | 0.83731 | 0.75042 |

In addition to compare the top-performing groups of models among the eight cycles of grid searches, we are also interested in knowing the performance of grid searches in identifying "decent" prediction models. An AUC of 0.5 is often treated as the worst prediction performance score, because it indicates that a model's prediction capability is equivalent to that of a random guess. What we mean by a "decent" prediction model is a model that scores at least as high as what we call a *mid-point score*, which is the average of 0.5 and the highest score of all models trained during the corresponding grid searches. So, a "decent" prediction model found in a cycle is a model that has an AUC no less than the mid-point score of the cycle of grid searches. A mid-point group is the group of "decent" prediction models found in a cycle of grid searches. Table 7 below compares the results concerning the mid-point groups of the eight cycles of grid searches.

**Table 7:** Mid-point group results of grid searches

Best-mean: the highest mean_test_AUC of all corresponding models; Mid-point : the average of 0.5 and the best-mean AUC; CHS: the count of hyperparameter settings at which the mean_test_AUCs are at least as high as the corresponding mid-point AUC; TNS: the total number of hyperparameter settings tested during the corresponding cycle of grid searches; CHS/TNS: the ratio of CHS over TNS; Avg-mean: the average mean_test_AUC of all models no worse than the mid-point AUC; Y: year.

|  |  | Stage1 | Stage2 | Stage3-c1 | Stage3-c2 | Stage3-c3 | Stage3-c4 | Stage3-c5 | Stage3-c6 |
|---|---|---|---|---|---|---|---|---|---|
| 5Y | Best-mean | 0.75301 | 0.74812 | 0.75024 | 0.75066 | 0.75418 | 0.75386 | 0.75250 | 0.75825 |
|  | Mid-point | 0.62650 | 0.62406 | 0.62512 | 0.62533 | 0.62709 | 0.62693 | 0.62625 | 0.62913 |
|  | CHS | 39112 | 510494 | 203991 | 101250 | 202500 | 404726 | 408625 | 43035 |
|  | TNS | 52042 | 582471 | 202500 | 101250 | 202500 | 405000 | 409050 | 70568 |
|  | CHS/TNS | 0.75155 | 0.87643 | 0.99946 | 1.0 | 1.0 | 0.99932 | 0.99896 | 0.60984 |
|  | Avg-mean | 0.71116 | 0.70278 | 0.72102 | 0.73505 | 0.73657 | 0.72059 | 0.72010 | 0.70415 |
| 10Y | Best_mean | 0.77789 | 0.78280 | 0.78802 | 0.79023 | 0.79152 | 0.78885 | 0.78059 | 0.78564 |
|  | Mid_point | 0.63895 | 0.64140 | 0.64401 | 0.64512 | 0.64576 | 0.64442 | 0.64029 | 0.64282 |
|  | CHS | 48517 | 635770 | 202500 | 60750 | 202495 | 405000 | 404994 | 83896 |
|  | TNS | 52416 | 638244 | 202500 | 60750 | 202500 | 405000 | 405000 | 105500 |
|  | CHS/TNS | 0.92561 | 0.99612 | 1.0 | 1.0 | 0.99998 | 1.0 | 0.99999 | 0.79522 |
|  | Avg_mean | 0.72116 | 0.73400 | 0.76124 | 0.75901 | 0.76702 | 0.76194 | 0.74337 | 0.72642 |
| 15Y | Best_mean | 0.86502 | 0.86548 | 0.88023 | 0.87463 | 0.87359 | 0.87551 | 0.87255 | 0.86657 |
|  | Mid_point | 0.68251 | 0.68274 | 0.69011 | 0.68731 | 0.68679 | 0.68776 | 0.68628 | 0.68328 |
|  | CHS | 41065 | 952677 | 337492 | 162000 | 270000 | 2058750 | 405000 | 73007 |
|  | TNS | 52090 | 1048575 | 337500 | 162000 | 270000 | 2058750 | 405000 | 93000 |
|  | CHS/TNS | 0.78835 | 0.90854 | 0.99998 | 1.0 | 1.0 | 1.0 | 1.0 | 0.78502 |
|  | Avg_mean | 0.800615 | 0.80289 | 0.83167 | 0.84535 | 0.83399 | 0.83782 | 0.83731 | 0.81619 |

As previously mentioned, managing the running time of a grid search is a challenging task. Knowing the RTPS, the average time it takes to train models at one hyperparameter setting, can be very useful in various aspects such as pre-estimating the total running time of a grid search, and estimating number of hyperparameter settings that a grid search take in when time we have to run a grid search is limited. During each of the eight cycles of grid searches, we periodically updated the corresponding RTPS based on the results we already obtained at the time, and then use it to guide upcoming experiments. Table 8 below shows the final RTPSs in seconds, the total number of hyperparameter settings (TNS), and the total running time (TRT) in hours for each cycle of grid searches we conducted, corresponding to the 5-year, 10-year, and 15-year models each respectively.

**Table 8:** Running time and count of hyperparameter settings concerning grid searches

RTPS: running time per hyperparameter setting in seconds; TNS: total number of hyperparameter settings; TRT: total running time in hours.

| Stage | 5-year | | | 10-year | | | 15-year | | |
|---|---|---|---|---|---|---|---|---|---|
| | RTPS | TNS | TRT | RTPS | TNS | TRT | RTPS | TNS | TRT |
| Stage1 | 42.42 | 52042 | 613.23 | 26.57 | 52416 | 386.87 | 20.06 | 52090 | 290.25 |
| Stage2 | 5.65 | 582471 | 914.72 | 3.36 | 638244 | 594.86 | 2.49 | 1048575 | 725.06 |
| Stage3-c1 | 10.65 | 202500 | 599.01 | 3.55 | 202500 | 199.42 | 2.43 | 337500 | 227.39 |
| Stage3-c2 | 11.97 | 101250 | 336.54 | 5.49 | 60750 | 92.66 | 1.63 | 162000 | 73.17 |
| Stage3-c3 | 11.11 | 202500 | 624.88 | 6.50 | 202500 | 365.77 | 1.25 | 270000 | 93.60 |
| Stage3-c4 | 25.46 | 405000 | 2863.73 | 7.66 | 405000 | 861.75 | 1.89 | 2058750 | 1080.01 |
| Stage3-c5 | 66.80 | 409050 | 7590.07 | 29.44 | 405000 | 3311.76 | 12.22 | 398252 | 1352.36 |
| Stage3-c6 | 116.40 | 70568 | 2281.77 | 51.05 | 105500 | 1495.91 | 27.36 | 93000 | 706.71 |

Figures 2-4 each contains eight SHAP feature importance plots generated using the eight best DFNN models, each obtained from one of the eight cycles of grid searches, for predicting 5-year, 10-year, and 15-year breast cancer metastasis respectively. In such a plot, each feature's mean absolute SHAP value is shown on the Y-axis, and the feature names are shown on the X-axis. Figure 5 contains two types of SHAP plots, the SHAP heatmaps (Figure 5(a), 5(c), and 5(e) ) and SHAP summary plots (Figure 5 (b), 5(d), and 5(f)), produced using the three best DFNN models for predicting 5-year, 10-year, and 15-year breast cancer metastasis, each respectively. Each of the three best Models is selected out of all models trained in the eight cycles of grid searches. A SHAP heatmap details the SHAP value distribution for all features, allowing the revelations of value patterns of grouped cases and features. In such a plot, features are ordered by their SHAP feature importance values from high to low as shown by a black bar plot to the right of the heatmap, and cases are grouped by the similarity of their SHAP impact based on hierarchical clustering. The notation $f(x)$ represents the distribution of the model's predictions. The heatmap plots as shown in Figure 5 were produced using the default hierarchical clustering method included in the SHAP package [41]. In companion with a SHAP heatmap, a SHAP summary plot not only shows the ordered average impact of all features, but also shows the specific SHAP values of the test cases with respect to each value of a feature. Therefore, such a plot reflects how a value of a feature influences the prediction. The summary plots in Figure 5 were also generated using the SHAP package.

We further analyze and demonstrate the interpretation of a model's prediction using the SHAP dependence plots with interaction visualization, as shown in Figure 6. To generate each of the subfigures in Figure 6, we used a best model found from all models trained during the corresponding eight cycles of grid searches. Figure 6(a), (c), and (e) are the correlation bar plots for the 5-year, 10-year, and 15-year predictions each respectively, which rank the association strength between each feature and the SHAP values of the most important feature identified using the SHAP feature importance. Using Figure 6(e), the figure for the 15-year prediction as an example, since the most import feature based on the SHAP feature importance value (see Figure 5(e) or 5(f)) is AGE (*age_at_diagonosis*) , Figure 6(e) shows that the feature that is mostly correlated with the SHAP impact of AGE is LYS (*lymph_node_status*). Figure 6(b), (d), and (f) are the corresponding dependence plots with interaction visualization, which demonstrate the interaction between the most import feature and the feature that is mostly correlated with the impact of the most important feature. Again using the 15-year as an example, since LYS is mostly correlated with the SHAP impact of the most important feature AGE, as shown in Figure 6(e), Figure 6(f) visualizes the joint effect of AGE and LYS. In Figure 6(f), with respect to each individual case, represented as a colored dot, the value of the most important feature, AGE, is plotted on the X axis, the SHAP value for the most important feature is shown on the Y-axis, and the color shows the value of LYS, the feature that is mostly correlated with the SHAP impact of AGE.

In the stage3-c6 grid searches, we used the RGS strategy (see Methods), which allows us to use a broad range of values for each of the hyperparameters (see Table S4-S6), therefore a large number of different hyperparameter settings were used in these grid searches. Specifically, in this cycle, 70,568, 105,500, and 93,000 different hyperparameter settings were used to train models for predicting 5-year, 10-year, and 15-year breast cancer metastasis, each respectively. Since the information about each hyperparameter setting and the corresponding mean_test_AUC from the 5-fold CV process are both recorded in our grid-search output, we were able to conduct SHAP analyses using these output as new data to identify the important hyperparameters to the predicted mean_test_AUC scores. In these SHAP analyses, each of the DFNN model hyperparameters was treated as a predictor (a feature), and the mean_test_AUC was treated as the outcome variable.

Our procedure for conducting this unique application of the SHAP analyses is as follows: 1) Process the grid search results so that they can be used as the input dataset of the SHAP analyses. For example, we recoded free-text like values of a categorical hyperparameter such as *kinit* (*kernel initializer*) with digits. Using kinit as an example, we converted the value *constant* to 0, *glorot_normal* to 1, *glorot_uniform* to 2, *he_normal* to 3, and *he_uniform* to 4; 2) Use the k-means algorithm to cluster the input dataset into n clusters, n is corresponding to the number of hyperparameters. Then use the centroid of each cluster to serve as the background data required by the SHAP Kernel Explainer (see the Methods); 3) Train a prediction model using 80% of the input data; For our analysis specifically, we trained a prediction model using the Random Forest method with 100 estimators; 4) Calculate the SHAP values using the 20% saved data and the prediction model obtained in 3), and then compute SHAP importance values for the features (the hyperparameters).

The results of the SHAP analyses concerning DFNN model hyperparameters are shown in Figure 7 below. Figure 7 (a), (c), and (e) are the SHAP feature importance plots concerning the hyperparameters, generated using the DFNN model grid-search output for predicting 5-year, 10-year, and 15-year breast cancer metastasis, each respectively, and Figure 7 (b), (d), and (f) are the corresponding SHAP summary plots.

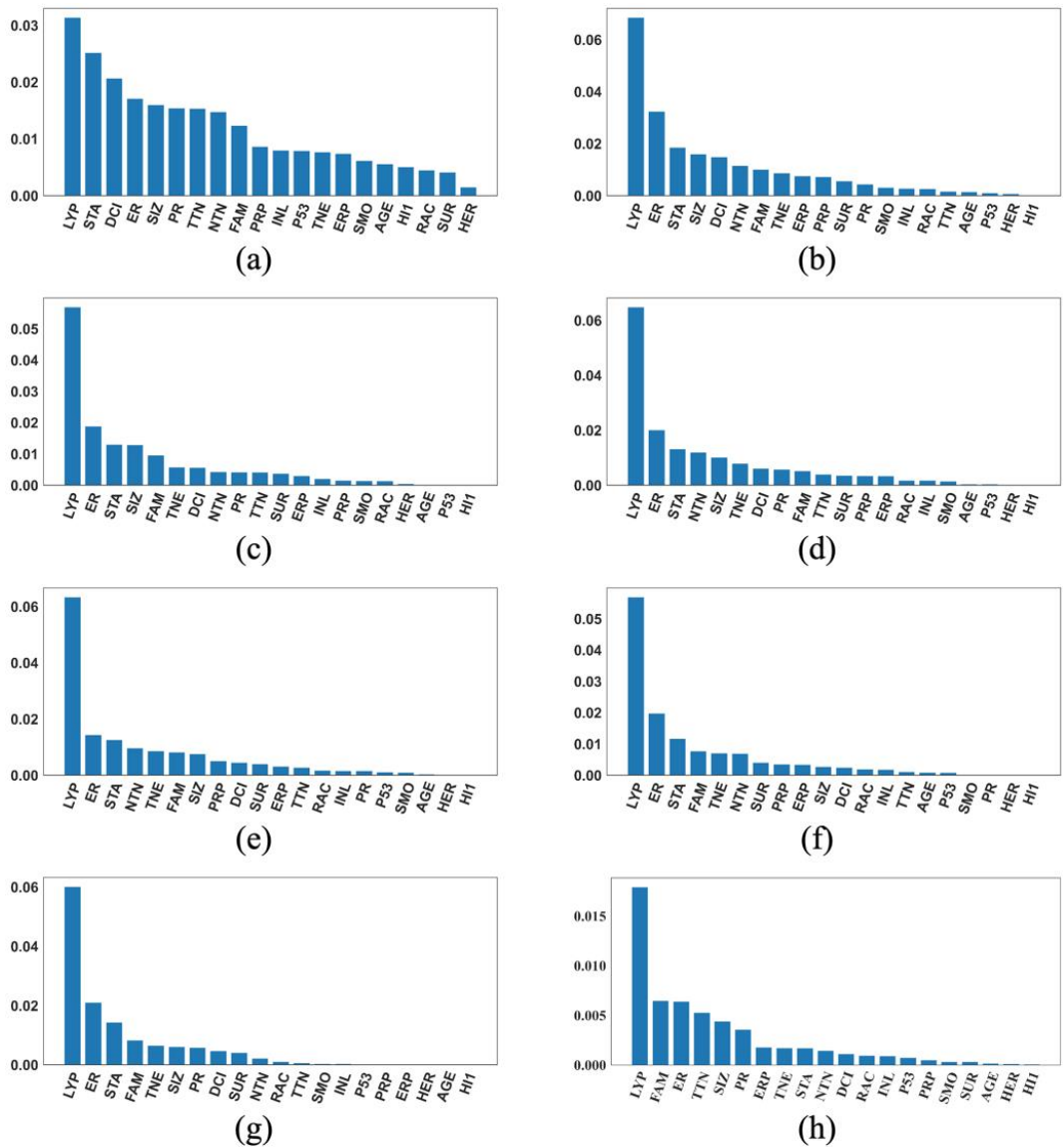

**Figure 2.** Feature importance plots of the best cycle-wise DFNN-5Year models
**AGE**: age at diagnosis of the disease; **ALC**: alcohol usage; **DCI**: type of ductal carcinoma in situ; **ER**: estrogen receptor expression; **ERP**: percent of cell stain pos for ER receptors; **ETH**: ethnicity; **FAM**: family history of cancer; **GRA**: grade of disease; **HER**: HER2 expression; **HI1**: tumor histology**;** **HI2**: tumor histology subtypes; **INL**: where invasive tumor is located; **INV**: whether tumor is invasive; **LYP**: number of positive lymph nodes; **LYR**: number of lymph nodes removed; **LYS**: patient had any positive lymph nodes; **MEN**: inferred menopausal status; **MRI**: MRIs within 60 days of surgery**;** **NTN**: number of nearby cancerous lymph nodes; **PR**: progesterone receptor expression; **PRP**: percent of cell stain pos for PR receptors; **P53**: whether P53 is mutated; **RAC**: race; **REE**: removal of an additional margin of tissue; **SID**: side of tumor; **SIZ**: size of tumor in mm; **SMO**: smoking; **STA**: composite of size and # positive nodes; **SUR**: whether residual tumor; **TNE**: triple negative status in terms of patient being ER, PR, and HER2 negative; **TTN**: prime tumor stage in TNM system.

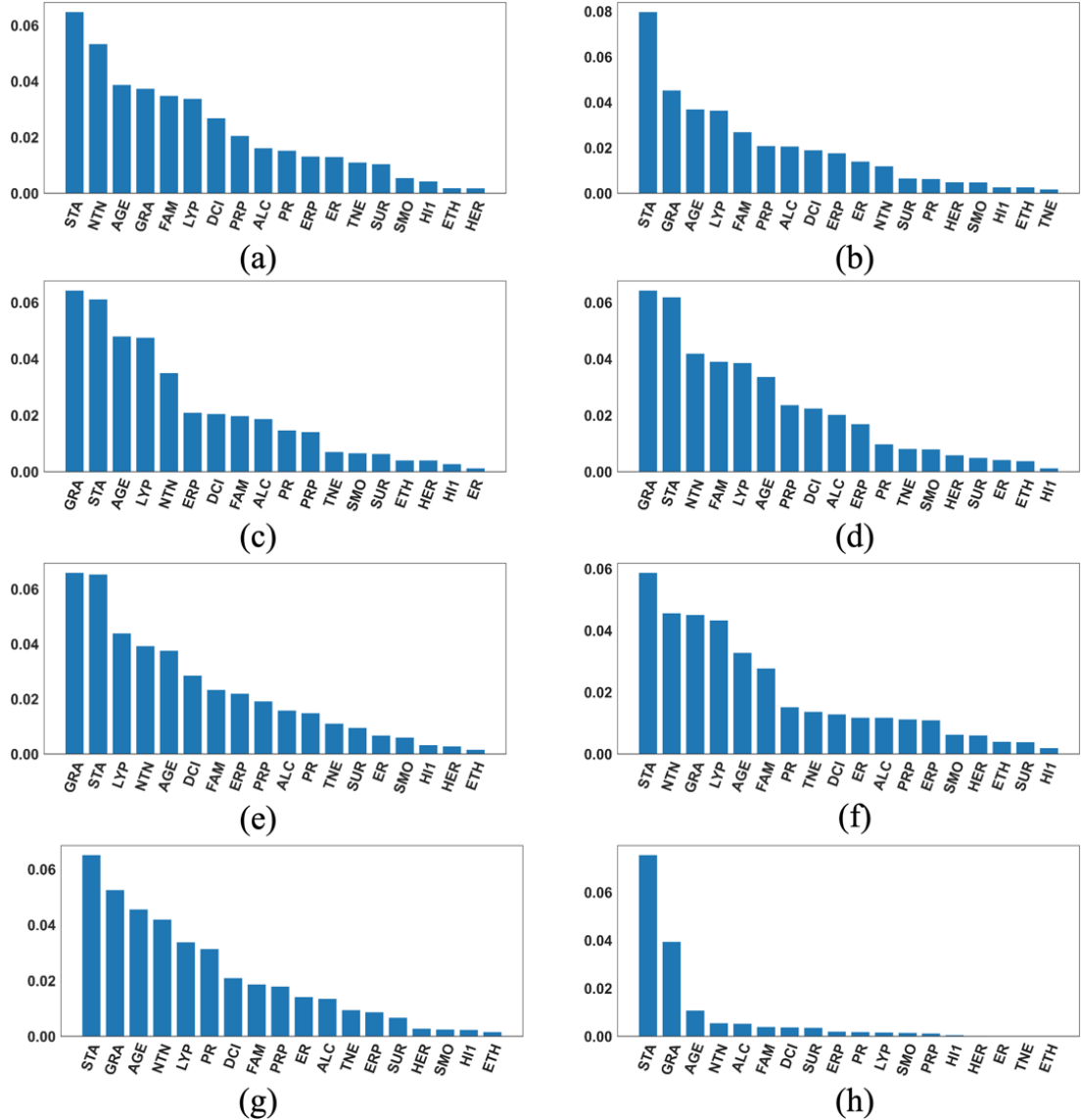

**Figure 3.** Feature importance plots of the best cycle-wise DFNN-10Year models
**AGE**: age at diagnosis of the disease; **ALC**: alcohol usage; **DCI**: type of ductal carcinoma in situ; **ER**: estrogen receptor expression; **ERP**: percent of cell stain pos for ER receptors; **ETH**: ethnicity; **FAM**: family history of cancer; **GRA**: grade of disease; **HER**: HER2 expression; **HI1**: tumor histology**; HI2**: tumor histology subtypes; **INL**: where invasive tumor is located; **INV**: whether tumor is invasive; **LYP**: number of positive lymph nodes; **LYR**: number of lymph nodes removed; **LYS**: patient had any positive lymph nodes; **MEN**: inferred menopausal status; **MRI**: MRIs within 60 days of surgery**; NTN**: number of nearby cancerous lymph nodes; **PR**: progesterone receptor expression; **PRP**: percent of cell stain pos for PR receptors; **P53**: whether P53 is mutated; **RAC**: race; **REE**: removal of an additional margin of tissue; **SID**: side of tumor; **SIZ**: size of tumor in mm; **SMO**: smoking; **STA**: composite of size and # positive nodes; **SUR**: whether residual tumor; **TNE**: triple negative status in terms of patient being ER, PR, and HER2 negative; **TTN**: prime tumor stage in TNM system.

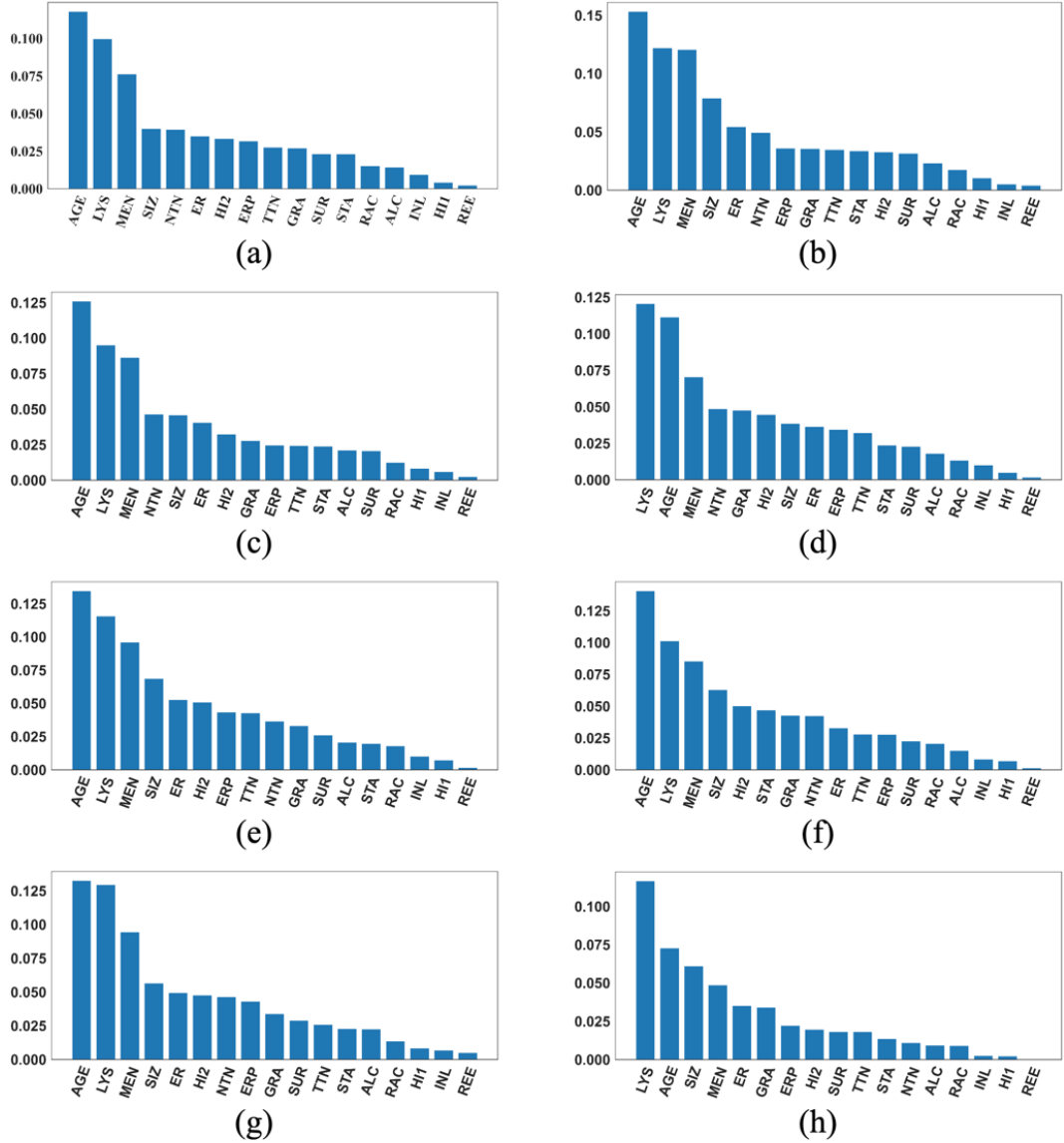

**Figure 4.** Feature importance plots of the best cycle-wise **D**FNN-15Year models
**AGE**: age at diagnosis of the disease; **ALC**: alcohol usage; **DCI**: type of ductal carcinoma in situ; **ER**: estrogen receptor expression; **ERP**: percent of cell stain pos for ER receptors; **ETH**: ethnicity; **FAM**: family history of cancer; **GRA**: grade of disease; **HER**: HER2 expression; **HI1**: tumor histology; **HI2**: tumor histology subtypes; **INL**: where invasive tumor is located; **INV**: whether tumor is invasive; **LYP**: number of positive lymph nodes; **LYR**: number of lymph nodes removed; **LYS**: patient had any positive lymph nodes; **MEN**: inferred menopausal status; **MRI**: MRIs within 60 days of surgery; **NTN**: number of nearby cancerous lymph nodes; **PR**: progesterone receptor expression; **PRP**: percent of cell stain pos for PR receptors; **P53**: whether P53 is mutated; **RAC**: race; **REE**: removal of an additional margin of tissue; **SID**: side of tumor; **SIZ**: size of tumor in mm; **SMO**: smoking; **STA**: composite of size and # positive nodes; **SUR**: whether residual tumor; **TNE**: triple negative status in terms of patient being ER, PR, and HER2 negative; **TTN**: prime tumor stage in TNM system.

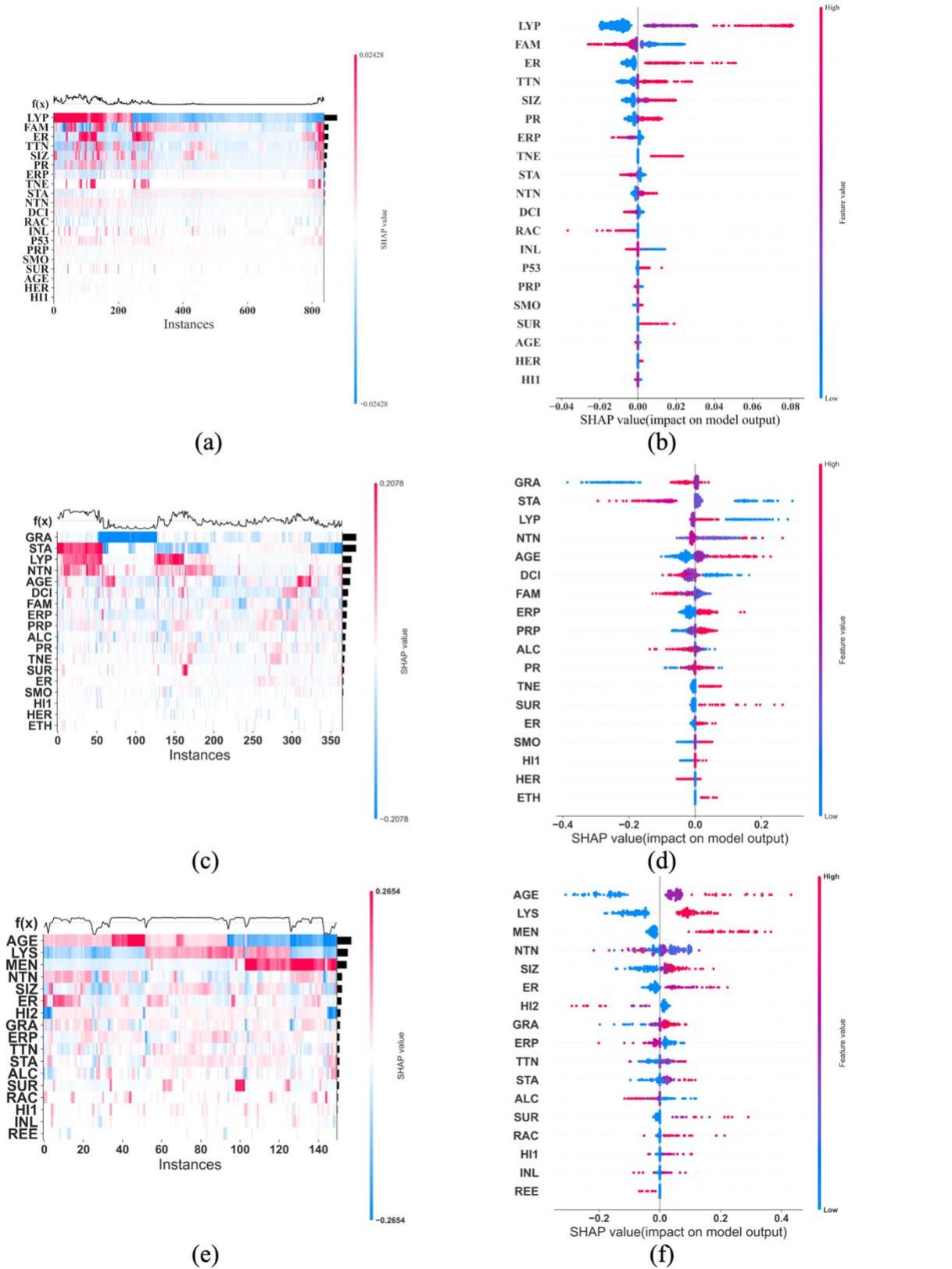

**Figure 5.** SHAP heatmap and summary plot for the best DFNN-5Year, 10Year, and 15Year model respectively

**AGE**: age at diagnosis of the disease; **ALC**: alcohol usage; **DCI**: type of ductal carcinoma in situ; **ER**: estrogen receptor expression; **ERP**: percent of cell stain pos for ER receptors; **ETH**: ethnicity; **FAM**: family history of cancer; **GRA**: grade of disease; **HER**: HER2 expression; **HI1**: tumor histology; **HI2**: tumor histology subtypes; **INL**: where invasive tumor is located; **INV**: whether tumor is invasive; **LYP**: number of positive lymph nodes; **LYR**: number of lymph nodes removed; **LYS**: patient had any positive lymph nodes; **MEN**: inferred menopausal status; **MRI**: MRIs within 60 days of surgery; **NTN**: number of nearby cancerous lymph nodes; **PR**: progesterone receptor expression; **PRP**: percent of cell stain pos for PR receptors; **P53**: whether P53 is mutated; **RAC**: race; **REE**: removal of an additional margin of tissue; **SID**: side of tumor; **SIZ**: size of tumor in mm; **SMO**: smoking; **STA**: composite of size and # positive nodes; **SUR**: whether residual tumor; **TNE**: triple negative status in terms of patient being ER, PR, and HER2 negative; **TTN**: prime tumor stage in TNM system.

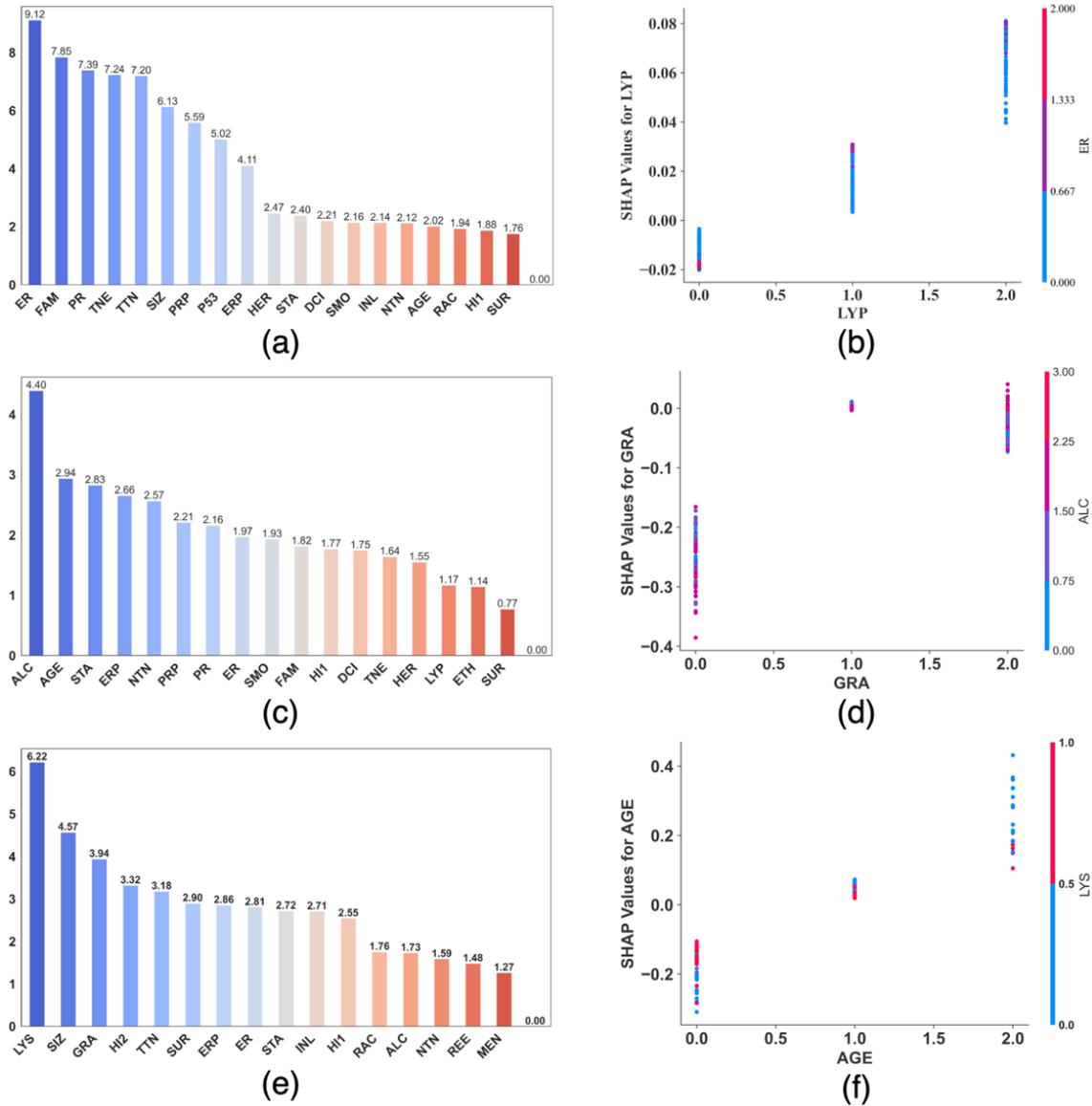

**Figure 6.** Correlation bar plot and independence plot with interaction visualization for the best DFNN-5Year, DFNN-10Year, and DFNN-15Year model, each respectively

**AGE**: age at diagnosis of the disease; **ALC**: alcohol usage; **DCI**: type of ductal carcinoma in situ; **ER**: estrogen receptor expression; **ERP**: percent of cell stain pos for ER receptors; **ETH**: ethnicity; **FAM**: family history of cancer; **GRA**: grade of disease; **HER**: HER2 expression; **HI1**: tumor histology; **HI2**: tumor histology subtypes; **INL**: where invasive tumor is located; **INV**: whether tumor is invasive; **LYP**: number of positive lymph nodes; **LYR**: number of lymph nodes removed; **LYS**: patient had any positive lymph nodes; **MEN**: inferred menopausal status; **MRI**: MRIs within 60 days of surgery; **NTN**: number of nearby cancerous lymph nodes; **PR**: progesterone receptor expression; **PRP**: percent of cell stain pos for PR receptors; **P53**: whether P53 is mutated; **RAC**: race; **REE**: removal of an additional margin of tissue; **SID**: side of tumor; **SIZ**: size of tumor in mm; **SMO**: smoking; **STA**: composite of size and # positive nodes; **SUR**: whether residual tumor; **TNE**: triple negative status in terms of patient being ER, PR, and HER2 negative; **TTN**: prime tumor stage in TNM system.

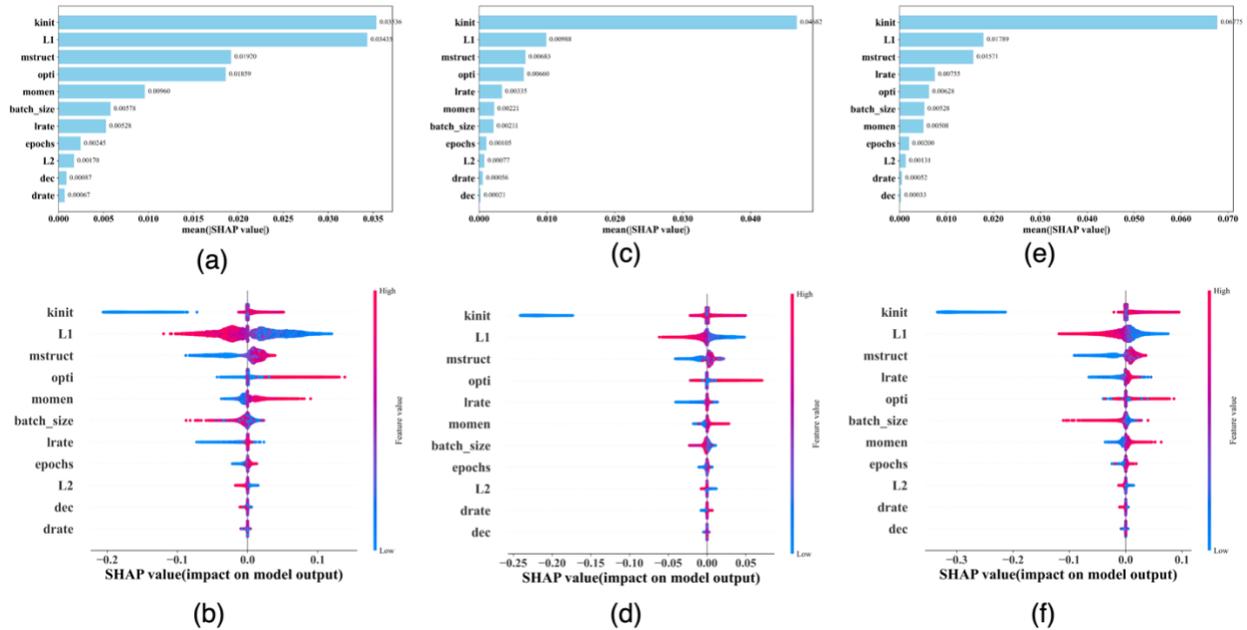

**Figure 7.** The SHAP Importance and summary plots concerning important hyperparameters for explaining predicted mean_test_AUCs based on results of DFNN-5Year, 10-Year, and 15-Year models respectively

**AGE**: age at diagnosis of the disease; **ALC**: alcohol usage; **DCI**: type of ductal carcinoma in situ; **ER**: estrogen receptor expression; **ERP**: percent of cell stain pos for ER receptors; **ETH**: ethnicity; **FAM**: family history of cancer; **GRA**: grade of disease; **HER**: HER2 expression; **HI1**: tumor histology**; HI2**: tumor histology subtypes; **INL**: where invasive tumor is located; **INV**: whether tumor is invasive; **LYP**: number of positive lymph nodes; **LYR**: number of lymph nodes removed; **LYS**: patient had any positive lymph nodes; **MEN**: inferred menopausal status; **MRI**: MRIs within 60 days of surgery**; NTN**: number of nearby cancerous lymph nodes; **PR**: progesterone receptor expression; **PRP**: percent of cell stain pos for PR receptors; **P53**: whether P53 is mutated; **RAC**: race; **REE**: removal of an additional margin of tissue; **SID**: side of tumor; **SIZ**: size of tumor in mm; **SMO**: smoking; **STA**: composite of size and # positive nodes; **SUR**: whether residual tumor; **TNE**: triple negative status in terms of patient being ER, PR, and HER2 negative; **TTN**: prime tumor stage in TNM system.

## DISCUSSION

Based on Table 4, for predicting 5-year breast cancer metastasis, the best performing model comes from Stage3-c6, with a mean_test_AUC that is higher than the average mean_test_AUC of all models trained in this cycle by 18.6%. Since we applied the RGS strategy in the Stage3-c6 grid searches, the average mean_test_AUC of all models trained in this cycle reflects the expected model performance when a hyperparameter value is randomly selected from a set of proper values. Table 5 shows that for predicting 10-year breast cancer metastasis, the best performing model comes also from stage 3 but is in cycle 3, with a mean_test_AUC that is higher than the average mean_test_AUC of all models trained in Stage3-c6 by 16.3%. For predicting 15-year breast cancer metastasis, the best performing model is found in stage3-c1, with a mean_test_AUC that is higher than the average mean_test_AUC of all models trained in Stage3-c6 by 17.3%.

Table 4-6 also show that the three-stage grid search mechanism overall works as expected due o the following reasons: Firstly, all best models are found in the Stage 3 grid searches. This is consistent with our goal of further refining grid searches in Stage 3 by building upon the preparation work from the Stage 1 and Stage 2 grid searches; Secondly, we conducted SSGSs in Stage3-c1 through Stage3-c3 grid searches, focusing on refining prediction perform based on the "sweet spot" derived from the results of its previous cycle. As expected, both Table 4 and 5 show a steady increase in terms of both the best and group-average model performance across these three cycles of grid searches; And thirdly, we used the RGS strategy in Stage3-c6, in which we allowed a hyperparameter to have a chance of taking any value in its proper set of values, determined based on Stage 1 grid searches. We reckon that such a grid search may accidently identify a best model because it has more "freedom" in choosing its hyperparameter value combinations, but its group model performance may not be as good because the pure randomness in selecting hyperparameter values can also result in very bad models. This indeed is reflected in our results as shown in Table 4 through Table 6.

By Table 4-6 we not only compare the prediction performance of the best models, but also compare group model performance among the five groups including top 5, 10, 50, 100, and "All". The group performance comparisons reflect somewhat the characteristics of the grid searches of the different stages, in which different strategies were applied. As seen in Table 4 through Table 6, Stage 1 grid searches tend to have a low "All" model performance. This is perhaps because in Stage 1 we ran grid searches that focused on one hyperparameter at a time, and therefore we were able to test a broad range of hyperparameter values. The extreme values used for a hyperparameter could lead to very bad models, which help drag down the average score of all models. Stage 2 overall does better than Stage 1 in group model performance, this is perhaps because the hyperparameter values used in stage 2 were selected from the set of proper values resulted from Stage 1. Table 4-6 also show that Stage 3 Cycle 1 through Cycle 5 tend to do better than both Stage 1 and Stage 2, and the explanation mostly lies in the fact that we conducted SSGSs in these cycles, in which the hyperparameter values used are "in the neighborhood" of some best values found in previous grid searches. The results seem to suggest that, in terms of model performance, if one model does well, then it's neighbors tend to do well also. In addition, Cycle 4 and 5 tend to do worse than Cycle 1 through 3 perhaps due to the OLO strategy. Our results also show that Stage 3 Cycle 6 grid searches are competitive in terms of group model performance except for the "All" model group. This indicates that the RGS strategy is competitive in terms of identifying top performing models, but the average performance of all models is brought down by performance outliers resulted from the pure randomness of hyperparameter value selection.

We trained hundreds of thousands of models at each of the eight cycles (see Table 8, and recall that 5 models were trained at each hyperparameter setting due to the 5-fold CV). Out of such a large number of models, how many of them are decent prediction models? An answer to this question may help further characterize the performance of grid searches. Table 7 shows the summary data that we derived from the output of our grid searches to answer this question. We described what we meant by a decent model in the Result section. Based on the CHS/TNS ratios in Table 7, Stage 3 Cycle 1 through Cycle 5 grid searches, that is, the so called SSGSs, did the best in terms of searching for a decent prediction model. The models trained by each of these five cycles are 100% or close to 100% decent, and this is true regardless which dataset was used. We also notice that the Stage 3 Cycle 6 grid searches produced the worst results in terms of identifying decent models, perhaps also due to the randomness of the hyperparameter value selection, required by the RGS strategy.

As described in the Introduction, time management has always been a critical issue in grid searches, especially in low-budget ones. Based on Table 8, Stage 1 and the SSGSs in Stage 3, especially the Cycle 1 through Cycle 4 tend to link to a relatively low RTPS, while the Stage 3 Cycle 6 for which we used the RGS strategy and Stage 2 tend to give a relatively high RTPS. By consulting Table S4 through S6, we found that the values of a hyperparameter called *the number of hidden nodes in a hidden layer* seem to positively correlate with the RTPS, because the value changes of this hyperparameter can well explain the changes of RTPS. For example, the highest RTPS that we see comes from Stage 3 Cycle 6, which can be explained by the range of high values (550-800) of this hyperparameter used in this cycle; Another example, we notice that Stage 1 allows this hyperparameter to take the highest value 1005, but only has the second highest RTPS. This can be explained by the broadest range of values (1 to 1005) that this hyperparameter takes in Stage 1, which should give an average value that is lower than the values used in Stage 3 Cycle 6. From Table 8, we also notice that within a same cycle of grid searches, where a same search strategy was used, the RTPSs for the three different datasets are also significantly different. The RTPS is the highest for the 5-year dataset, and lowest for the 15-year dataset. Based on Table 2, a significant difference among the three datasets is the size of a dataset, that is, the number of cases (also called data points in machine learning) contained in a dataset. The 5-year dataset contains the largest number of cases, while the 15-year contains the lowest number. Table 8 and Table 2 together reveal a positive correlation between the number of data points contained in a dataset and the RTPS of a grid search.

Based on Figure 2-4, the important features in the best models are quite consistent among the eight cycles. A variable called LYP (*lymph_nodes_positive*) is found to be the most import feature for predicting 5-year breast cancer metastasis in all eight best models shown in Figure 2. Besides, Figure 2 also demonstrates that LYP is far more important than the other features, and this is consistently demonstrated by the SHAP importance plots of almost all the best models. According to our knowledge about cancer, it seems to make sense that the number of positive lymph nodes is a good indicator for breast cancer metastasis at diagnosis, and therefore a good indicator for the risk of cancer recurrence in the near future. Based on Figure 2, the top three important features for predicting 5-year breast cancer metastasis also include ER (*estrogen_receptor_expression*) and STA (*stage*), with ER to be the second most important feature voted by five out of the eight best models, and STA to be the third most important feature voted by six out of the eight best models.

According to Figure 3, the top three most important features for predicting the risk of 10-year breast cancer include STA, identified as the most important in five out of the eight best models, GRA (*grade*), identified as the most important by three out of the eight best models, and AGE (*age_at_diagnosis*), identified as the number three important by five out of the eight best models. Unlike the situation where one feature mostly dominates as seen in the 5-year DFNN models, we see that more than one feature dominate in most of the importance plots. Based on Figure 4, the top three important features for predicting the risk of 15-year breast cancer metastasis are AGE, LYS (*lymph_nodes_status*) and MEN (*menopausal_status*). We also notice that in seven out of the eight best models, these three top features together dominate the overall predicted risk of breast cancer metastasis.

Although the feature importance figures (Figure 2-4) identify the top feature(s) that dominate(s) the overall importance of the predicted results, they don't show the relationships among the features in terms of their impact. The SHAP heatmaps as shown in Figure 5 perhaps do better in this regard. For example, Figure 5(e), a SHAP heatmap plot, reveals color patterns in terms of both features and cases, which indicate similar influences of the grouped members on the model's prediction. However, the heatmap plots don't show whether the features interact to have a joint effect on prediction.

An interactive effect is the additional effect after subtracting the individual effect from each of the interactive features. A SHAP dependence plot with interaction visualization such as the ones shown in Figure 6 carries more information in terms of a joint contribution between two features. For example, Figure 6 (e) shows that LYS is strongly correlated with the impact from AGE, which is the most important feature for predicting the risk of 15-year breast cancer metastasis (see Figure 5(e) and 5(f)). Figure 6(f) demonstrates that even at the same value of AGE, the SHAP values distribute differently at different values of LYS, confirming that AGE and LYS, the top two most influential features for predicting the risk of 15-year breast cancer metastasis, indeed interact to contribute to the prediction. More specifically, based on Figure 6(f), when the value AGE is 0 (for patients who are 0 to 49 years old), it has overall a negative impact on the predicted risk, when the value of AGE is 1 (for patients who are 50 to 69 years old), it has an overall positive but low impact on the predicted risk, when AGE is 2 (for patients who are above 70 years old), it has a significantly increased positive impact on the predicted risk. In addition, Figure 6(f) reveals that at low ages, LYS tends to be positively correlated with the SHAP impact of AGE, while as age increases, LYS tends to be negatively correlated with the SHAP impact of AGE.

As shown by Figure 7 (a), (c), and (e), the hyperparameter that has the highest overall impact on the predicted mean_test_AUC is kinit, which is used to assign the initial weights of some regression functions used in gradient descent, a key process in deep learning. This is a bit out of our expectation, because we originally thought for a deep learning, the depth of the network, which is reflected in a hyperparameter called *mstruct* should be the most important feature. The importance of kinit perhaps comes from its influences on gradient descent. An inappropriate value assignment of kinit can cause divergent learning, which should have a negative impact on model performance. The summary plots as shown in Figure 7 (b), (d), and (f), reveal that the value of kinit is positively correlated with its SHAP value, which seems to suggest that he_uniform, the highest value of kinit, has the highest positive impact on the predicted mean_test_AUC of a DFNN model. This may help explain that he_uniform was often associated with a good DFNN model, as seen in some preliminary experiments that we conducted [4].

Figure 7 also demonstrates that the second most important feature in terms of the predicted mean_test_AUC is L1, a coefficient of the LASSO regularization, which can help select features and therefore control model sparsity during training [43]. The summary plots in Figure 7 suggest a negative correlation between the values of L1 and its SHAP values. Mstruct ranks number three in terms of its importance to the predicted mean_test_AUCs. It is a two-dimensional hyperparameter for our DFNN models, consisting of two hyperparameters: *the number of hidden*

*layers* and the number of hidden nodes in a hidden layer. The summary plots in Figure 7 reveal a positive correlation between mstruct and its SHAP values, indicating that the complexity of a neural network structure may have a positive impact on the predicted mean_test_AUCs. This may help explain the observed success of deep neural networks in contrast to the "shallow" neural networks, namely, the first-generation one-hidden layer neural networks. Finally, we also notice from Figure 7 that the rankings of the top three most important features are quite consistence in all plots. This may indicate that the most important adjustable hyperparameters for the DFNN models are not so much dependent on a dataset. However, since this is the first time, as far as we know, that a SHAP analysis is used to explain the role of a hyperparameter as a feature in a predicted mean_test_AUC, the implications of the kind of results shown in Figure 7 are subjected to further explorations.

# CONCLUSIONS

Our grid searches improve the risk prediction of 5-year, 10-year, and 15-year breast cancer metastasis by 18.6%, 16.3%, and 17.3%, relatively to the average mean_test_AUC of all corresponding models, for which the value of each hyperparameter was randomly picked from a set of reasonable values. Our result analyses characterize grid searches in terms of their capabilities of identifying groups of top performers and "decent" models. The heuristic three-stage mechanism worked effectively. It enables us to finish the low-budget grid searches within expected timeframe. The results of the grid searches are consistent with what we expected, that is, all the best performing models are found in Stage 3, the final model-refining stage. The local cycles of SSGSs in Stage 3 overall show steadily increased performance not only in terms of the best models, but also in different groups of models. The SSGS Strategy significantly beats the RGS strategy in terms of the average performance of all models and percentage of "decent" models found in all models, but the RGS is quite competitive with the SSGS in identifying best models. The unit grid search time is positively correlated with both the number of hidden nodes in a hidden layer and the number of data points contained in a dataset. The results also suggest that a model with a set of hyperparameter values "close to" the set of values of a good prediction model tends to be a good model. The SHAP analyses not only reveal the features with high impact on the predicted risk of breast cancer metastasis, but also help identify pairs of interactive features in regards to SHAP values. Finally, an unique SHAP application demonstrates that the kernel initializer, L1, and mstruct are the top three most important hyperparameters to the predicted mean_test_AUCs.

# DECLARATIONS

## Ethics approval and consent to participate

The study was approved by University of Pittsburgh Institutional Review Board (IRB # 196003) and the U.S. Army Human Research Protection Office (HRPO # E01058.1a).

The need for patient consent was waived by the ethics committees because the data consists only of de-identified data that are publicly available.

## Consent for publication

Not applicable.

## Availability of data and material

The data used in this study are available at datadryad.org (DOI 10. 5061/dryad.64964m0).

## Competing interests

The authors declare that they have no competing interests.

## Funding

Research reported in this paper was supported by the U.S. Department of Defense through the Breast Cancer Research Program under Award No. W81XWH-19-1-0495 (to XJ). Other than supplying funds, the funding agencies played no role in the research.

## Authors' Contribution

XJ originated the study and designed the methods. XJ and YZ wrote the first draft of the manuscript. XJ, CX, and YZ implemented the methods. CX and YZ conducted the experiments. XJ, YZ, CX prepared and analyzed the results. All authors contributed to the preparation and revision of the manuscript. All work was conducted in the University of Pittsburgh.

# Supplement

**Table S1**: Predictors in the LSM_RF-5Year Dataset

|    | Predictors | Description | Values |
|----|---|---|---|
| 1  | race | race of patient | white, black, Asian, American Indian or Alaskan native, native Hawaiian or other Pacific islander |
| 2  | smoking | smoking history of patient | ex smoker, non smoker, cigarettes, chewing tobacco, cigar |
| 3  | family history | family history of cancer | cancer, no cancer, breast cancer, other cancer, cancer but nos |
| 4  | age_at_diagnosis | age at diagnosis of the disease | 0-49, 50-69, >69 |
| 5  | TNEG | triple negative status in terms of patient being ER, PR, and HER2 negative | yes, no |
| 6  | ER | estrogen receptor expression | neg, pos, low pos |
| 7  | ER_percent | percent of cell stain pos for ER receptors | 0-20, 20-90, 90-100 |
| 8  | PR | progesterone receptor expression | neg, pos, low pos |
| 9  | PR_percent | percent of cell stain pos for PR receptors | 0-20, 20-90, 90-100 |
| 10 | P53 | P53 | whether P53 is mutated |
| 11 | HER2 | HER2 expression | neg, pos |
| 12 | t_tnm_stage | prime tumor stage in TNM system | 0, 1, 2, 3, 4, IS, 1mic, X |
| 13 | n_tnm_stage | # of nearby cancerous lymph nodes | 0, 1, 2, 3, 4, X |
| 14 | stage | composite of size and # positive nodes | 0, 1, 2, 3 |
| 15 | lymph_nodes_positive | number of positive lymph nodes | 0, 1-8, >8 |
| 16 | histology | tumor histology | lobular, duct |
| 17 | size | size of tumor in mm | 0-32, 32-70, >70 |
| 18 | invasive_tumor_location | where invasive tumor is located | mixed duct and lobular, duct, lobular, none |
| 19 | DCIS_level | type of ductal carcinoma in situ | solid, apocrine, cribriform, dcis, comedo, papillary, micropapillary |
| 20 | surgical_margins | whether residual tumor | res. tumor, no res. tumor, no primary site surgery |

**Table S2**. Predictors in the LSM_RF-10Year Dataset

|    | Predictors | Description | Values |
|----|---|---|---|
| 1 | ethnicity | ethnicity of patient | not Hispanic, Hispanic |
| 2 | smoking | smoking history of patient | ex smoker, non smoker, cigarettes, chewing tobacco, cigar |
| 3 | alcohol usage | alcohol usage of patient | moderate, no use, use but nos (non otherwise specified), former user, heavy user |
| 4 | family history | family history of cancer | cancer, no cancer, breast cancer, other cancer, cancer but nos |
| 5 | age_at_diagnosis | age at diagnosis of the disease | 0-49, 50-69, >69 |
| 6 | TNEG | triple negative status in terms of patient being ER, PR, and HER2 negative | yes, no |
| 7 | ER | estrogen receptor expression | neg, pos, low pos |
| 8 | ER_percent | percent of cell stain pos for ER receptors | 0-20, 20-90, 90-100 |
| 9 | PR | progesterone receptor expression | neg, pos, low pos |
| 10 | PR_percent | percent of cell stain pos for PR receptors | 0-20, 20-90, 90-100 |
| 11 | HER2 | HER2 expression | neg, pos |
| 12 | n_tnm_stage | # of nearby cancerous lymph nodes | 0, 1, 2, 3, 4, X |
| 13 | stage | composite of size and # positive nodes | 0, 1, 2, 3 |
| 14 | lymph_nodes_positive | number of positive lymph nodes | 0, 1-8, >8 |
| 15 | histology | tumor histology | lobular, duct |
| 16 | grade | grade of disease | 1, 2, 3 |
| 17 | DCIS_level | type of ductal carcinoma in situ | solid, apocrine, cribriform, dcis, comedo, papillary, micropapillary |
| 18 | surgical_margins | whether residual tumor | res. tumor, no res. tumor, no primary site surgery |

**Table S3**: Predictors in the LSM_RF-15 Year Dataset

| | Predictors | Description | Values |
|---|---|---|---|
| 1 | race | race of patient | white, black, Asian, American Indian or Alaskan native, native Hawaiian or other Pacific islander |
| 2 | alcohol usage | alcohol usage of patient | moderate, no use, use but nos (non otherwise specified), former user, heavy user |
| 3 | age_at_diagnosis | age at diagnosis of the disease | 0-49, 50-69, >69 |
| 4 | menopausal_status | inferred menopausal status | pre, post |
| 5 | ER | estrogen receptor expression | neg, pos, low pos |
| 6 | ER_percent | percent of cell stain pos for ER receptors | 0-20, 20-90, 90-100 |
| 7 | t_tnm_stage | prime tumor stage in TNM system | 0, 1, 2, 3, 4, IS, 1mic, X |
| 8 | n_tnm_stage | # of nearby cancerous lymph nodes | 0, 1, 2, 3, 4, X |
| 9 | stage | composite of size and # positive nodes | 0, 1, 2, 3 |
| 10 | lymph_node_status | patient had any positive lymph nodes | neg, pos |
| 11 | size | size of tumor in mm | 0-32, 32-70, >70 |

| 12 | grade | grade of disease | 1, 2, 3 |
| 13 | histology2 | tumor histology subtypes | IDC, DCIS, ILC, NC |
| 14 | invasive_tumor_location | where invasive tumor is located | mixed duct and lobular, duct, lobular, none |
| 15 | re_excision | removal of an additional margin of tissue | yes, no |
| 16 | surgical_margins | whether residual tumor | res. tumor, no res. tumor, no primary site surgery |
| 17 | histology | tumor histology | lobular, duct |

**Table S4.** Range of hyperparameter values used in the three-stage grid searches for predicting 5 year breast cancer metastasis

| Hyperparameter Name | Hyperparameter Values | | | | | | | |
|---|---|---|---|---|---|---|---|---|
| | Stage 1 | Stage 2 | Stage 3-c1 | Stage 3-c2 | Stage 3-c3 | Stage 3-c4 | Stage 3-c5 | Stage 3-c6 |
| # of Hidden Layers | 1,2,3,4 | 1,2,3,4 | 2,3,4 | 4 | 2,3,4 | 2,3,4 | 2,3,4 | 1,2,3,4 |
| # of Hidden Nodes Each Layer | 1 to 1005 | 1 to 350 | 100 to 350 | 100 to 200 | 100 to 350 | 200 to 350 | 350 to 550 | 550 to 800 |
| Activation Function | 'relu' | 'relu' | 'relu' | 'relu' | 'relu' | 'relu' | 'relu' | 'relu' |
| Kernel initializer | 'he_normal', 'glorot_normal' | 'Glorot_normal' | 'Glorot_normal' | 'Glorot_normal' | 'Glorot_normal' | 'Glorot_normal' | 'Glorot_normal' | 'Constant', 'Glorot_normal', 'Glorot_uniform', 'He_normal', 'He_uniform' |
| Optimizer | 'SGD', 'Adagrad' | 'Adagrad' | 'Adagrad' | 'Adagrad' | 'Adagrad' | 'Adagrad' | 'Adagrad' | 'SGD','Adagrad','adam' |
| Learning rate | 0.001 to 0.3 | 0.01 to 0.1 | 0.01 to 0.09 | 0.05 to 0.09 | 0.07 to 0.09 | 0.01 to 0.09 | 0.01 to 0.09 | 0.01 to 0.09 |
| Momentum | 0 to 0.9 | 0 to 0.4 | 0.1 to 0.4 | 0.1 to 0.2 | 0.12 to 0.18 | 0.1 to 0.4 | 0.1 to 0.4 | 0.1 to 0.9 |
| Iteration-based Decay | 0 to 0.1 | 0 to 0.001 | 0 to 0.0005 | 0 to 0.0006 | 0.0004 to 0.0006 | 0 to 0.0005 | 0 to 0.1 | 0 to 0.001 |
| Dropout rate | 0 to 0.5 | 0 to 0.2 | 0 to 0.1 | 0 to 0.05 | 0.03 to 0.05 | 0 to 0.1 | 0 to 0.1 | 0 to 0.1 |
| Epochs | 5 to 2000 | 25 to 175 | 100 to 180 | 140 to 180 | 160 to 180 | 100 to 180 | 100 to 180 | 100 to 180 |
| Batch_size | 1 to 4189 | 100 to 1000 | 100 to 500 | 300 to 540 | 460 to 540 | 200 to 500 | 100 to 500 | 100 to 500 |
| L1 | 0 to 0.03 | 0 to 0.01 | 0 to 0.01 | 0.004 to 0.006 | 0.004 to 0.005 | 0 to 0.01 | 0 to 0.01 | 0 to 0.03 |
| L2 | 0 to 0.2 | 0 to 0.05 | 0 to 0.01 | 0 to 0.01 | 0 to 0.005 | 0 | 0 | 0 to 0.03 |

**Table S5.** Range of hyperparameter values used in the three-stage grid searches for predicting 10 year breast cancer metastasis

| Hyperparameter Name | Hyperparameter Values | | | | | | | |
|---|---|---|---|---|---|---|---|---|
| | Stage 1 | Stage 2 | Stage 3-c1 | Stage 3-c2 | Stage 3-c3 | Stage 3-c4 | Stage 3-c5 | Stage 3-c6 |
| # of Hidden Layers | 1,2,3,4 | 1,2,3,4 | 2,3,4 | 4 | 2,3,4 | 2,3,4 | 2,3,4 | 1,2,3,4 |
| # of Hidden Nodes Each Layer | 5 to 1005 | 50 to 200 | 50 to 200 | 50 to 150 | 50 to 200 | 50 to 200 | 350 to 550 | 550 to 800 |
| Activation Function | 'relu' | 'relu' | 'relu' | 'relu' | 'relu' | 'relu' | 'relu' | 'relu' |
| Kernel initializer | 'he_normal', 'glorot_normal' | 'Glorot_normal' | 'Glorot_normal' | 'Glorot_normal' | 'Glorot_normal' | 'Glorot_normal' | 'Glorot_normal' | 'Constant', 'Glorot_normal', 'Glorot_uniform', 'He_normal', 'He_uniform' |
| Optimizer | 'SGD', 'Adagrad' | 'Adagrad' | 'Adagrad' | 'Adagrad' | 'Adagrad' | 'Adagrad' | 'Adagrad' | 'SGD','Adagrad','adam' |
| Learning rate | 0.001 to 0.299 | 0.01 to 0.2 | 0.01 to 0.1 | 0.03 to 0.05 | 0.025 to 0.29 | 0.01 to 0.1 | 0.01 to 0.09 | 0.01 to 0.09 |
| Momentum | 0 to 0.9 | 0.4 | 0.1 to 0.4 | 0.1 to 0.2 | 0.1 to 0.15 | 0.1 to 0.4 | 0.1 to 0.4 | 0.1 to 0.9 |
| Iteration-based Decay | 0 to 0.01 | 0 to 0.001 | 0 to 0.001 | 0.0006 to 0.001 | 0.0009 to 0.0011 | 0 to 0.001 | 0 to 0.0005 | 0 to 0.001 |
| Dropout rate | 0 to 0.5 | 0 to 0.2 | 0.1 to 0.2 | 0.1 to 0.25 | 0.21 to 0.24 | 0.1 to 0.2 | 0 to 0.1 | 0 to 0.1 |
| Epochs | 5 to 1960 | 10 to 160 | 110 to 150 | 150 to 190 | 165 to 185 | 110 to 150 | 100 to 180 | 100 to 180 |
| Batch_size | 1 to 1827 | 40 to 250 | 40 to 250 | 40 to 120 | 30 to 90 | 40 to 250 | 100 to 500 | 100 to 500 |
| L1 | 0 to 0.199 | 0 to 0.01 | 0.002 to 0.005 | 0 to 0.003 | 0.002 to 0.004 | 0.002 to 0.005 | 0 to 0.01 | 0 to 0.03 |
| L2 | 0 to 0.199 | 0 to 0.02 | 0 to 0.005 | 0 to 0.008 | 0 to 0.003 | 0 | 0 | 0 to 0.03 |

**Table S6.** Range of hyperparameter values used in the three-stage grid searches for predicting 15 year breast cancer metastasis

| Hyperparameter Name | Hyperparameter Values | | | | | | | |
|---|---|---|---|---|---|---|---|---|
| | Stage 1 | Stage 2 | Stage 3-c1 | Stage 3-c2 | Stage 3-c3 | Stage 3-c4 | Stage 3-c5 | Stage 3-c6 |
| # of Hidden Layers | 1,2,3,4 | 1,2,3,4 | 3,4 | 3,4 | 3,4 | 2,3,4 | 2,3,4 | 1,2,3,4 |
| # of Hidden Nodes Each Layer | 5 to 1005 | 50 to 150 | 50 to 110 | 50 to 110 | 50 to 110 | 50 to 140 | 350 to 550 | 550 to 800 |
| Activation Function | 'relu' | 'relu' | 'relu' | 'relu' | 'relu' | 'relu' | 'relu' | 'relu' |
| Kernel initializer | 'he_normal', 'glorot_normal' | 'Glorot_normal' | 'Glorot_normal' | 'Glorot_normal' | 'Glorot_normal' | 'Glorot_normal' | 'Glorot_normal' | 'Constant', 'Glorot_normal', 'Glorot_uniform', 'He_normal', 'He_uniform' |
| Optimizer | 'SGD', 'Adagrad' | 'Adagrad' | 'Adagrad' | 'Adagrad' | 'Adagrad' | 'Adagrad' | 'Adagrad' | 'SGD','Adagrad','adam' |
| Learning rate | 0.001 to 0.299 | 0.01 to 0.1 | 0.01 to 0.1 | 0.03 to 0.07 | 0.03 to 0.05 | 0.01 to 0.1 | 0.01 to 0.09 | 0.01 to 0.09 |
| Momentum | 0 to 0.9 | 0.4 | 0.1 to 0.4 | 0.3 to 0.5 | 0.35 to 0.45 | 0.1 to 0.4 | 0.1 to 0.4 | 0.1 to 0.9 |
| Iteration-based Decay | 0 to 0.01 | 0 to 0.001 | 0 to 0.001 | 0 to 0.0005 | 0 to 0.0002 | 0 to 0.001 | 0 to 0.0005 | 0 to 0.001 |
| Dropout rate | 0 to 0.5 | 0 to 0.2 | 0 to 0.2 | 0 to 0.1 | 0 to 0.05 | 0 to 0.2 | 0 to 0.1 | 0 to 0.1 |
| Epochs | 5 to 1960 | 5 to 95 | 50 to 95 | 50 to 90 | 70 to 110 | 50 to 95 | 100 to 180 | 100 to 180 |
| Batch_size | 1 to 751 | 60 to 200 | 60 to 200 | 300 to 540 | 50 to 110 | 60 to 200 | 100 to 500 | 100 to 500 |
| L1 | 0 to 0.199 | 0 to 0.025 | 0 to 0.002 | 0 to 0.003 | 0.00035 to 0.00045 | 0 to 0.002 | 0 to 0.01 | 0 to 0.03 |
| L2 | 0 to 0.199 | 0 to 0.025 | 0 to 0.025 | 0 to 0.005 | 0.0015 to 0.003 | 0 to 0.025 | 0 | 0 to 0.03 |